\documentclass[11pt]{article}

\usepackage[final]{acl}

\usepackage{times}
\usepackage{latexsym}
\usepackage{graphicx}
\usepackage{algorithm}
\usepackage{algpseudocode}
\usepackage{amsmath}

\usepackage[utf8]{inputenc}
\usepackage[T1]{fontenc}
\usepackage{url}
\usepackage{booktabs}
\usepackage{amsfonts}
\usepackage{amsmath}
\usepackage{amssymb}
\usepackage{nicefrac}
\usepackage{multirow}
\usepackage{tikz}
\usepackage{pgfplots}
\usepackage{xspace}
\usepackage{tabularx}
\usepackage{pifont}  %
\usepackage{listings}
\usepackage{stfloats}

\pgfplotsset{compat=1.18}

\definecolor{codekeyword}{rgb}{0.00,0.20,0.65}  %
\definecolor{codestring}{rgb}{0.55,0.10,0.55}   %
\definecolor{codecomment}{rgb}{0.20,0.50,0.20}  %
\lstdefinestyle{pythoncode}{
    basicstyle=\ttfamily\small,
    commentstyle=\color{codecomment}\itshape,
    keywordstyle=\color{codekeyword}\bfseries,
    stringstyle=\color{codestring},
    breaklines=true, breakatwhitespace=false,
    captionpos=b, keepspaces=true,
    numbers=left, numbersep=8pt, numberstyle=\tiny\color{gray},
    showstringspaces=false,
    frame=tb, framerule=0.5pt, framesep=4pt,
    xleftmargin=2.5em, aboveskip=8pt, belowskip=8pt,
    abovecaptionskip=4pt,
    language=Python,
}

\newcommand{\methodname}{VTOS}

\title{VTOS: Learning to Orchestrate Vision Tools by Co-Searching Solutions and Observers}
\author{
  Jinchao Ge \\
  University of Wollongong \\
  \texttt{jge@uow.edu.au} \\
  \And
  Lingqiao Liu\thanks{Corresponding author.} \\
  Adelaide University \\
  \texttt{lingqiao.liu@adelaide.edu.au} \\
  \AND
  Shuwen Zhao \\
  Tianjin University of Technology \\
  \texttt{DonutZsw@gmail.com} \\
  \And
  Lei Wang \\
  University of Wollongong \\
  \texttt{leiw@uow.edu.au} \\
}

\begin{document}
\maketitle

\begin{abstract}
Vision foundation tools such as open-vocabulary detectors, segmentation models, and post-processing operators are powerful building blocks for computer vision, but their effectiveness depends heavily on how they are orchestrated: which tools are used, in what order, with what parameters, and under what visual conditions. Existing visual-programming agents typically generate a fixed solution pipeline, making them brittle under dense objects, occlusion, small targets, and domain shift. We introduce VTOS (Vision Tools Orchestration Search), a framework for adaptive visual tool orchestration through joint solution--observer search. VTOS co-searches executable solution programs that compose vision tools such as Grounding DINO, SAM, NMS, and slice-and-detect, together with observer programs that diagnose candidate solutions, identify failure modes, and generate actionable feedback. These observations are accumulated in a shared VisionThoughts knowledge base to guide subsequent search. We evaluate VTOS through two case studies: dense object counting on LVIS-Count and zero-shot plant-disease segmentation on PlantSeg-OOD, which stress different orchestration challenges including threshold calibration, NMS, slicing, mask refinement, and domain generalization. Across both tasks, VTOS outperforms static tool pipelines and agentic visual-programming baselines, specifically in complex settings such as dense, occluded scenes and out-of-distribution segmentation where static pipelines leave measurable headroom, rather than in standard tasks where a single well-calibrated tool already approaches its ceiling.

\end{abstract}

\begin{figure*}[!t]
\centering

\includegraphics[width=\textwidth]{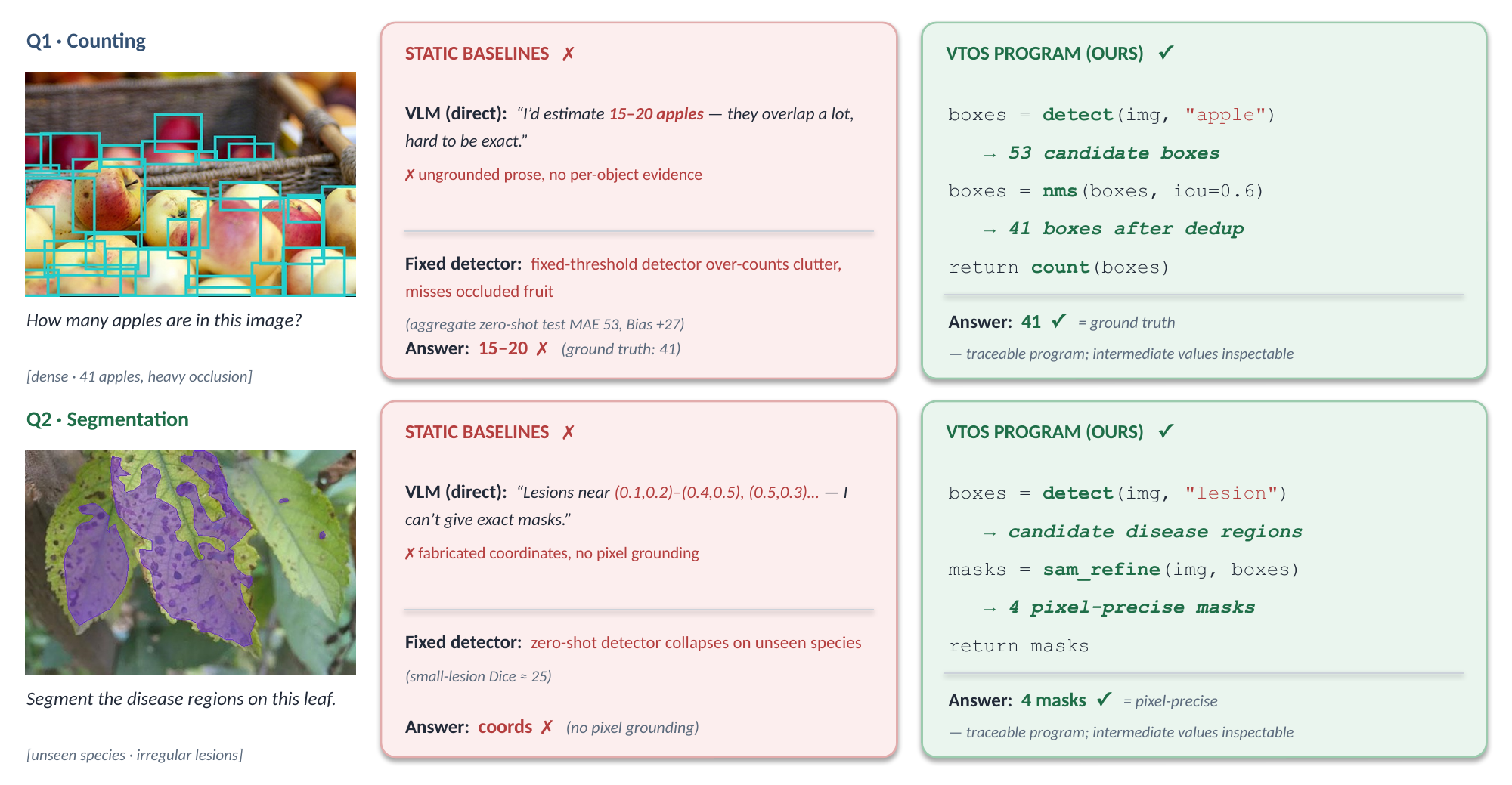}
\caption{\textbf{Static tool use fails on hard scenes; \methodname{} searches an auditable, image-adaptive program.}
For two tasks that stress different orchestration failures, dense object counting (Q1) and out-of-distribution leaf-disease segmentation (Q2), directly applying a VLM yields ungrounded prose answers, while a fixed-threshold detector breaks down under clutter, occlusion, and unseen species (aggregate zero-shot test MAE $53$ / Bias $+27$ for counting; small-lesion Dice ${\approx}\,25$ for segmentation).
The middle column groups these two search-free baselines. \methodname{} instead searches a short, executable program with auditable intermediate steps (e.g., candidate detections ${\to}$ NMS ${\to}$ final count) to yield the correct answer. The predicted bounding boxes and segmentation masks are overlaid on the input images (left). The two input panels show enlarged center-crops of the original images for visual clarity; reported counts correspond to the full images, so not all counted instances are visible within the crop.
The deployed artifact is an inspectable program, not opaque prose.}
\label{fig:concept_comparison}
\end{figure*}

\section{Introduction}

The development of vision foundation models has changed how computer vision systems can be built. Instead of collecting task-specific data and retraining a dedicated model for every new task, practitioners can increasingly solve vision problems by orchestrating reusable tools: open-vocabulary detectors can localize text-described concepts, segmentation models can refine spatial prompts into masks, visual-language models can provide semantic inspection, and classical operators can structure, filter, or aggregate intermediate outputs. This tool-based paradigm makes computer vision development more flexible and lightweight, especially when labeled data or training resources are limited. However, it also introduces a new challenge: performance now depends not only on the capability of each tool, but on how these tools are composed into an effective task-solving procedure\footnote{Code: \url{https://github.com/jinchaogjc/VTOS}}.

Visual tool orchestration is far from trivial. A task may require decomposing a high-level request into perceptual subproblems, selecting suitable tools, deciding how intermediate outputs should be represented and passed between tools, and invoking alternative strategies when an initial pipeline fails. Dense counting, for example, requires coordinating detection, duplicate handling, spatial aggregation, and density-aware processing, while fine-grained segmentation often requires localization, mask refinement, and post-processing to work together. As vision foundation tools become more general-purpose, the bottleneck increasingly shifts from training new models to discovering how to orchestrate existing tools.

While recent visual-programming methods (e.g., ViperGPT~\citep{suris2023vipergpt}, VisProg~\citep{gupta2023visual}) provide an important step toward tool orchestration, they operate in an open-loop fashion without systematically learning from execution failures. Therefore, we argue that effective use of vision foundation tools requires more than one-shot program synthesis. A useful agent should learn an \emph{orchestration algorithm}: an executable procedure that decomposes the task, composes available tools, manages intermediate outputs, and adapts its strategy when previous attempts fail.

This motivates the key idea of \textbf{\methodname{}} (Vision Tools Orchestration Search): \emph{co-searching solutions and observers}. In \methodname{}, the search process learns two coupled forms of executable knowledge. The first is a \emph{solution program}, which orchestrates vision tools to solve the target task. The second is an \emph{observer program}, which inspects candidate solutions, analyzes their execution behavior, and produces diagnostic observations that guide future solution generation. The observer is not a fixed critic prompt or a hand-designed evaluator; it is itself a searched program whose role is to transform execution feedback into actionable insight.

The framework realizes this idea through a Producer--Analyzer architecture. The Producer proposes candidate solution programs from a constrained tool library, and each candidate is executed and evaluated in a sandboxed environment. The Analyzer proposes observer programs that examine top-ranked solutions using execution traces, numerical summaries, visual diagnostics, and a budgeted set of visual-inspection tools. The resulting observations are accumulated in a shared \emph{VisionThoughts} Knowledge Base, which stores empirical insights, tool-use hypotheses, and exploration directions. This knowledge conditions subsequent rounds of both solution and observer search, forming a closed loop in which the system improves both its task-solving strategy and its understanding of previous failures. Ultimately, the final output of this search is an inspectable Python program that orchestrates vision foundation tools and can be deployed efficiently without invoking VLMs at test time.

We then evaluate \methodname{} through two case studies that expose different visual-tool orchestration challenges. The first is dense object counting on LVIS-Count, a 180-image subset sampled from LVIS, where successful solutions must coordinate detection, spatial aggregation, duplicate handling, and density-aware processing under object crowding and occlusion. The second is zero-shot plant-disease segmentation on PlantSeg-OOD, a 180-image species-disjoint subset sampled from PlantSeg, where open-vocabulary localization must be combined with mask refinement and post-processing under species-level domain shift and small target regions. Both tasks were chosen because static tool pipelines struggle on them: crowding and occlusion degrade fixed detection thresholds, while species-level domain shifts break fixed segmentation pipelines. They are not intended to represent computer vision at large, but rather settings where joint solution-observer search is most effective; for tasks where a single well-calibrated tool already saturates performance, additional search offers little gain.

In summary, our contributions are:
\begin{itemize}
    \item \textbf{Tool orchestration as algorithm search.}
    We formulate computer vision task solving as learning orchestration algorithms over reusable vision foundation tools, rather than retraining task-specific models.

    \item \textbf{Joint solution--observer co-search via a Producer--Analyzer loop.}
    We introduce a co-search framework that learns both task-solving programs and diagnostic observer programs, instantiated through a Producer--Analyzer architecture whose observations accumulate in a shared \emph{VisionThoughts} Knowledge Base to guide subsequent search.

    \item \textbf{Case studies on counting and segmentation.}
    We evaluate \methodname{} on LVIS-Count and PlantSeg-OOD, showing improvements over static tool pipelines and agentic visual-programming baselines on both dense object counting and out-of-distribution segmentation tasks.
\end{itemize}
\section{Related Work}
\label{sec:related}

\noindent \textbf{Code-Based Visual Reasoning.}
Neuro-symbolic approaches that compile natural-language queries into executable programs have emerged as a compelling alternative to monolithic vision-language models (VLMs), which are widely observed to hallucinate numerical and spatial details when asked to count objects or localize regions, the failure mode contrasted in Figure~\ref{fig:concept_comparison}. ViperGPT~\citep{suris2023vipergpt} and VisProg~\citep{gupta2023visual} synthesize Python programs invoking pre-trained vision APIs, yielding compositional interpretability that end-to-end models such as LLaVA~\citep{liu2023visual} cannot provide. AutoMMLab~\citep{yang2024autommlab} and Kim et al.~\citeyearpar{kim2025autonomous} extend this paradigm toward fully autonomous computer-vision pipelines driven by natural-language specifications. Despite their promise, these systems operate in an \textit{open-loop} fashion: they rely on a single generation pass and lack mechanisms to interpret execution failures or iteratively revise solutions. Program synthesis research~\citep{gao2023pal} has long recognized the importance of iterative refinement, yet the closed-loop integration of runtime feedback with a structured search over candidate algorithms remains underexplored. \methodname{} closes this gap by treating both runtime errors and analyzer-observed failure modes as informative signals that drive a Producer--Analyzer algorithm search, enabling systematic hypothesis revision rather than one-shot generation.

\noindent \textbf{LLM-Based Agentic Reasoning.}
ReAct~\citep{yao2023react} demonstrated the efficacy of alternating reasoning and action steps in language models, and PAL~\citep{gao2023pal} showed that program-aided decomposition grounds LLM reasoning in executable logic. MM-ReAct~\citep{yang2023mmreact} extends this paradigm to multimodal settings by coupling ChatGPT with vision-expert pools. Self-correction through feedback loops has gained traction via Reflexion~\citep{shinn2023reflexion}, which uses verbal reinforcement learning for iterative improvement, REFINER~\citep{paul2024refiner}, which trains a critic to provide structured reasoning feedback, and Self-Refine~\citep{madaan2023selfrefine}, which has a single LLM iteratively critique and refine its own output. The Producer--Analyzer split in \methodname{} inherits this generator/critic structural motif, which it shares with actor-critic and generator-discriminator architectures more broadly. Unlike those approaches, however, \methodname{} anchors the distinction in \textit{tool affordances} rather than prompts: the Analyzer is given a separate, budgeted Visual Toolbox that the Producer cannot access. Consequently, deployed programs cannot smuggle in analyzer-only VLM tools, and the asymmetric-modality boundary is preserved by construction. Unlike probabilistic self-critique or prompt-tuning approaches, \methodname{} grounds its correction signal in \textit{deterministic} interpreter execution. Syntax errors, tracebacks, and variable-state inspections provide ``hard'' constraints that enable more reliable search guidance.

\noindent \textbf{Inference-Time Visual Tool Use.}
A parallel line of work enables multimodal models to reason with visual tools dynamically at inference time. Visual Sketchpad~\citep{hu2024sketchpad} treats canvas drawings as visual chains of thought, while recent commercial systems integrate image manipulation directly into the reasoning loop~\citep{openai2025thinkingimages}. Similarly, recent multimodal agents extend tool capabilities through runtime API synthesis~\citep{marsili2025dynamicapi} or step-wise preference optimization~\citep{li2025itertool}. While expressive, these methods operate \emph{per query}: the VLM reasons over each image individually and discards the generated procedure once answered, incurring recurring VLM inference costs on every test sample. \methodname{} instead pays this cost once, offline, and deploys a program that runs without a VLM (\S\ref{sec:efficiency}).

\noindent \textbf{AutoML agent.}
Recent work has increasingly framed machine learning engineering as an LLM-agent-based search problem, where agents iteratively propose code, run experiments, observe validation feedback, and refine candidate solutions. MLE-Bench provides a key benchmark for this direction by evaluating agents on realistic Kaggle-style ML tasks involving data processing, model training, debugging, and submission generation \cite{chan2025mlebench}. Building on this benchmark, AIDE explores tree search over executable code solutions \cite{jiang2025aide}, R\&D-Agent decomposes the process into researcher and developer agents for iterative solution building \cite{yang2025rdagent}, and AI Research Agents for Machine Learning studies how different search policies, including greedy, evolutionary, and MCTS-style strategies, affect performance on MLE-Bench \cite{toledo2025ai}. Other recent systems improve specific components of the loop: MLE-STAR combines web-based model retrieval with targeted refinement \cite{nam2025mlestar}, MLE-Dojo provides an interactive Gym-style environment for training and evaluating ML agents \cite{qiang2025mledojo}, and recent work on ideation diversity and learned ideators highlights the importance of generating diverse and useful research directions before implementation \cite{audran-reiss2025what,zhang2026learning}. Synthetic Sandbox further studies scalable training environments for ML engineering agents \cite{zhou2026synthetic}. Together, these works move beyond classical AutoML over predefined model and hyperparameter spaces, toward open-ended search over ML programs, experimental strategies, and algorithmic design choices.
\section{Method}
\label{sec:method}

\begin{figure*}[!t]
\centering
\includegraphics[width=\textwidth]{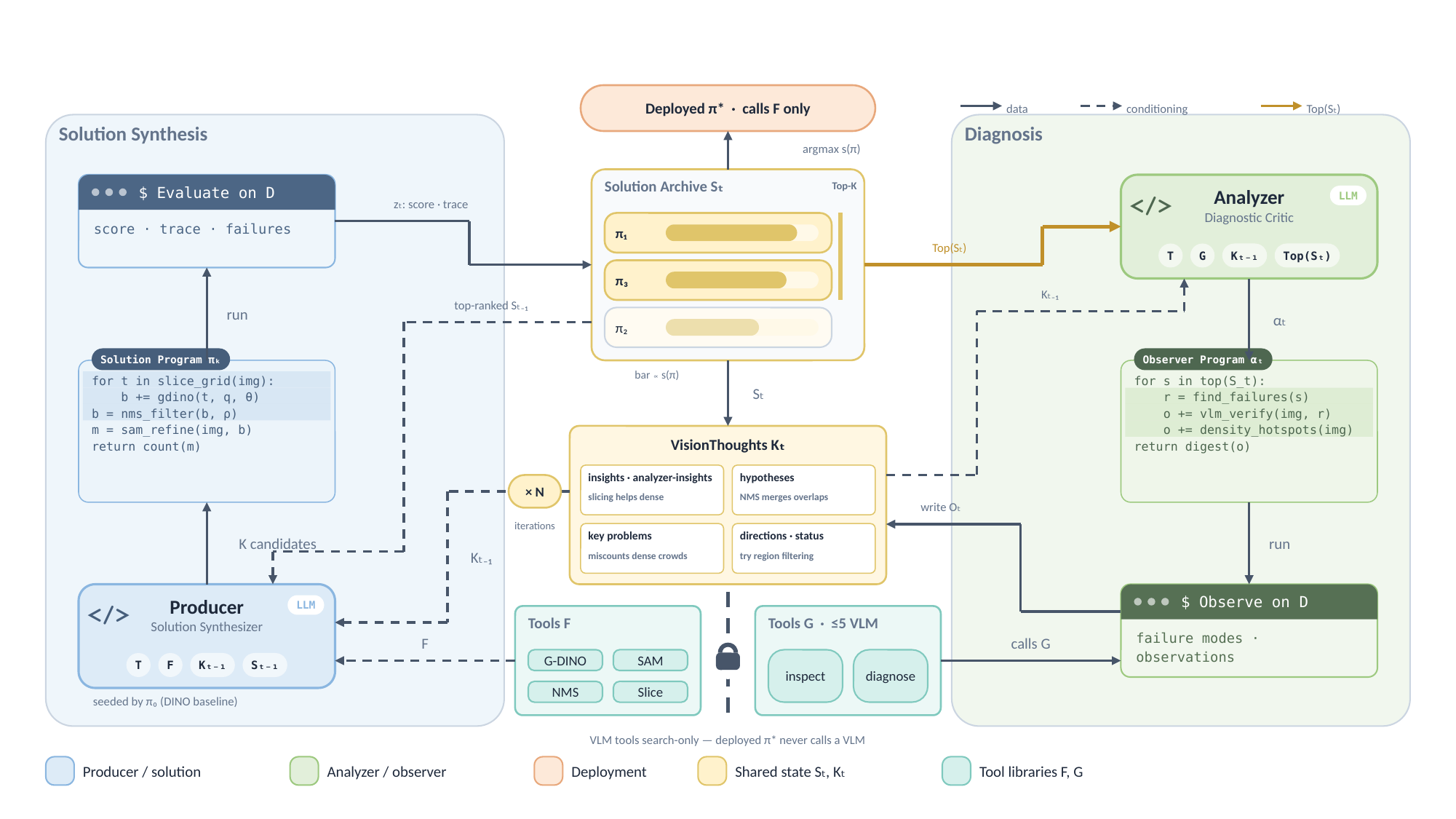}
\caption{\textbf{\methodname{} Framework.} The framework runs for ${\times}N$ iterations. In each iteration, a \textbf{Producer} proposes $K$ candidate solution programs from a constrained tool library; programs are executed and scored by a sandboxed subprocess evaluator. An \textbf{Analyzer} then proposes observer programs targeting the top-ranked solutions $\mathrm{Top}(\mathcal{S}_t)$ (amber coupling line), generating qualitative failure observations via a budgeted toolbox of pure-CV and VLM-as-eyes tools. All observations and empirical insights are written into a shared VisionThoughts Knowledge Base---the \emph{central} state carried across iterations. A dashed Search\,/\,Deployment Boundary separates the search phase from deployment: at test time only the top-ranked solution program is deployed (asymmetric modality---VLMs serve only as diagnostic feedback during search, not at inference).}
\label{fig:method_architecture}
\end{figure*}

We present \methodname{} (Vision Tools Orchestration Search), a framework for learning to orchestrate vision foundation tools through \emph{solution--observer co-search}. Figure~\ref{fig:method_architecture} illustrates the overall architecture. Instead of retraining model weights or generating a single open-loop program, \methodname{} searches over executable tool-use procedures for solving a target computer vision task. \methodname{} runs a single iterative search loop. In each iteration, a \textit{Producer} module (\S\ref{sec:producer}) first proposes candidate solution programs from a constrained tool library, and the system executes and evaluates these candidates to obtain scores, traces, intermediate outputs, and failure cases. An \textit{Analyzer} module (\S\ref{sec:analyzer}) then proposes observer programs for selected evaluated solutions, converting their execution behavior into diagnostic observations. These observations are written into a shared \textit{Knowledge Base} (\S\ref{sec:thoughts}), together with empirical insights, tool-use hypotheses, and exploration directions. The updated Knowledge Base conditions the next iteration, allowing future solution and observer proposals to be informed by previous execution feedback. Both the Producer and the Analyzer are instantiated by the \emph{same} underlying LLM, conditioned on different prompts and tool libraries; the same model also performs the Knowledge Base update. %

\subsection{Producer: History-Aware Search for Orchestration Programs}
\label{sec:producer}

At each iteration $t$, the Producer proposes candidate \emph{solution programs} that orchestrate the available vision tools to solve the target task. Each solution program $\pi$ is an executable Python procedure over a constrained solution-tool library $\mathcal{F}$, which may include open-vocabulary localization, segmentation, region processing, spatial aggregation, filtering, and post-processing primitives. The Producer is therefore not asked to answer a single visual query directly; rather, it searches for a reusable orchestration algorithm that can be applied across the task distribution.

The Producer is history-aware: it conditions on the task, the solution-tool library, prior solution snapshots, and the current Knowledge Base, making each proposal an informed revision rather than an independent restart.

Formally, the Producer generates $K$ candidate solution programs as
\begin{equation}
\begin{aligned}
    \Pi_t &= \{\pi^{(t)}_1,\ldots,\pi^{(t)}_K\} \\
    &\sim p_{\mathrm{LLM}}\bigl(\cdot \mid \mathcal{T}, \mathcal{F}, \mathcal{K}_{t-1}, \mathcal{S}_{t-1}\bigr),
\end{aligned}
\label{eq:producer}
\end{equation}
where $\mathcal{S}_{t-1}$ is the archive of evaluated solution snapshots. To constrain the context length, in practice we only select several top-ranked solutions from the archive $\mathcal{S}_{t-1}$ for the Producer to generate the next solutions. Information about other trials will be absorbed in the knowledge base $\mathcal{K}_{t-1}$.
The conditioning of $\mathcal{S}_{t-1}$ and $\mathcal{K}_{t-1}$ allows the Producer to generate candidates that are not independent restarts, but informed revisions of the current search trajectory. For example, if previous observers found that a pipeline failed because an intermediate representation was unreliable, the Producer can explore a different decomposition or introduce an alternative tool chain rather than merely perturbing the previous code. Such an agentic-style algorithm search has proven more effective as shown in recent studies \cite{qu2026coral}.

Each proposed solution program is executed in a sandboxed environment on the task dataset $\mathcal{D}$. The evaluator records the program output, task score, execution trace, errors, intermediate states, and representative success or failure cases:
\begin{equation}
\begin{aligned}
    z^{(t)}_k &= \mathrm{Evaluate}\bigl(\pi^{(t)}_k,\mathcal{D}\bigr), \\
    \mathcal{S}_t &= \mathcal{S}_{t-1} \cup \{z^{(t)}_k\}_{k=1}^{K}.
\end{aligned}
\label{eq:evaluate_solution}
\end{equation}
Here, $z^{(t)}_k$ denotes a solution snapshot containing both quantitative performance and execution evidence. Candidates are ordered by a Borda rank-sum $s(\pi)$: entries in the archive are ranked separately under each of the task's two metrics (mIoU and MAE for counting; Dice and mIoU for segmentation), with rank~$1$ assigned to the best performance in each metric's respective direction. Summing these ranks yields $s(\pi)$, where a smaller value is preferred and neither metric dominates. $\mathrm{Top}(\cdot)$ returns entries with the lowest rank-sums. The search operates exclusively on the training split; after $N$ iterations, the top candidates are re-evaluated on the held-out validation split, and the program with the lowest validation rank-sum is returned as $\pi^\star$ and deployed unchanged to the test set (Algorithm~\ref{alg:vas}). These snapshots are later inspected by the Analyzer and may also be selected as examples for future Producer calls. Constraining programs to $\mathcal{F}$ keeps the search executable and reproducible, while the history-aware prompt context lets the Producer explore different decompositions, tool compositions, control flows, intermediate representations, and recovery strategies.

\subsection{Analyzer: Learning Solution-Conditioned Diagnostics}
\label{sec:analyzer}

The Analyzer converts the behavior of candidate solutions into actionable feedback for the next search iteration. A task score can rank solution programs, but it rarely explains why a solution fails or how the orchestration should be revised. Therefore, \methodname{} treats diagnosis itself as a learnable object: at each iteration, the Analyzer generates an executable \emph{observer program} that inspects the current best solution attempts.

Formally, given the task $\mathcal{T}$, observer-tool library $\mathcal{G}$, Knowledge Base $\mathcal{K}_{t-1}$, and selected top solution snapshots $\mathrm{Top}(\mathcal{S}_t)$, the Analyzer proposes one observer program:
\begin{equation}
    \alpha_t \sim p_{\mathrm{LLM}}\bigl(\cdot \mid \mathcal{T}, \mathcal{G}, \mathcal{K}_{t-1}, \mathrm{Top}(\mathcal{S}_t)\bigr).
    \label{eq:analyzer}
\end{equation}
where $\alpha_t$ denotes the proposed observer program, generated by the same LLM as in Eq.~\eqref{eq:producer}. Conditioning on $\mathrm{Top}(\mathcal{S}_t)$ is important: it tells the LLM what the current best solutions are, how they are implemented, and where they still fail. The resulting diagnosis code is therefore customized to the actual solution strategies under consideration, rather than being a generic fixed critic.

Executing the observer program produces diagnostic observations:
\begin{equation}
    O_t = \alpha_t\bigl(\mathrm{Top}(\mathcal{S}_t), \mathcal{D};\, \mathcal{G}\bigr).
    \label{eq:observe}
\end{equation}
These observations may identify recurring failure modes, unreliable intermediate outputs, regime-specific weaknesses, or differences between competing solution strategies. They are then written into the Knowledge Base and used to guide future solution generation.

The observer program is used only during search; the final deployed solution is selected from the solution archive and calls only the solution-tool library $\mathcal{F}$, preserving an asymmetric-modality boundary at deployment.

\subsection{VisionThoughts: Knowledge-Guided Co-Search}
\label{sec:thoughts}

The Knowledge Base, denoted by $\mathcal{K}_t$, is the shared memory that connects iterations of solution--observer co-search. It stores what the system has learned from previous solution attempts and diagnostic observations, so that later generations are informed by the accumulated search history rather than starting from scratch.

$\mathcal{K}_t$ is organized into structured fields: insights, hypotheses, key problems, and exploration directions --- defined fully in Appendix~\ref{sec:appendix_kb_fields}.

After each iteration, the same LLM updates the Knowledge Base from the evaluated solution snapshots and the observer-generated diagnostic observations, deciding which insights and hypotheses to add, which to confirm or refute against the new evidence, and which exploration directions to pursue or prune:
\begin{equation}
    \mathcal{K}_t
    =
    \mathrm{Update}
    \bigl(
    \mathcal{K}_{t-1},
    \mathcal{S}_t,
    O_t
    \bigr).
    \label{eq:kb_update}
\end{equation}
This update records which orchestration strategies worked, which failed, under what conditions they failed, and what should be explored next. The updated $\mathcal{K}_t$ then conditions both the Producer and the Analyzer in the next iteration: the Producer uses it to generate better solution programs, while the Analyzer uses it to design more targeted diagnostic observers.

\begin{algorithm}[t]
\caption{\methodname{}: Solution--Observer Co-Search}
\label{alg:vas}
\begin{algorithmic}[1]
\Require Task $\mathcal{T}$, dataset $\mathcal{D}$, solution tools $\mathcal{F}$, observer tools $\mathcal{G}$, iterations $N$
\Ensure Best solution program $\pi^\star$
\State Initialize knowledge base $\mathcal{K}_0$ and solution archive $\mathcal{S}_0$
\For{$t = 1,\ldots,N$}
    \State Sample $K$ candidates: $\Pi_t \leftarrow \mathrm{Producer}(\mathcal{T}, \mathcal{F}, \mathcal{K}_{t-1}, \mathcal{S}_{t-1})$
    \State $\mathcal{S}_t \leftarrow \mathcal{S}_{t-1} \cup \mathrm{Evaluate}(\Pi_t, \mathcal{D})$
    \State $\alpha_t \leftarrow \mathrm{Analyzer}(\mathcal{T}, \mathcal{G}, \mathcal{K}_{t-1}, \mathrm{Top}(\mathcal{S}_t))$
    \State $O_t \leftarrow \mathrm{Observe}(\alpha_t, \mathrm{Top}(\mathcal{S}_t), \mathcal{D})$
    \State $\mathcal{K}_t \leftarrow \mathrm{Update}(\mathcal{K}_{t-1}, \mathcal{S}_t, O_t)$
\EndFor
\State \Return $\displaystyle \pi^\star = \arg\min_{\pi \in \mathrm{Top}(\mathcal{S}_N)} s_{\mathrm{val}}(\pi)$
\end{algorithmic}
\end{algorithm}

\begin{table*}[t]
\centering
\small
\caption{\textbf{LVIS-Count Overall Metrics.} Aggregated performance on the 100-image zero-shot test set (no density stratification). Bias is signed mean count error (negative = undercount; zero is best). Best per column in \textbf{bold}. All \methodname{} rows use Sonnet 4.6 as the LLM and Grounding DINO + SAM as the Vision Foundation Model (VFM) toolbox. Per-density breakdown is in Table~\ref{tab:sota_per_density}.}
\label{tab:sota_comparison}
\resizebox{0.8\textwidth}{!}{%
\begin{tabular}{@{}llcccc@{}}
\toprule
\textbf{Family} & \textbf{Method} & MAE$\downarrow$ & RMSE$\downarrow$ & Bias ($\downarrow |\cdot|$) & mIoU\%$\uparrow$ \\
\midrule
\textit{End-to-End VLM} & Sonnet 4.6 (Count Only) & 35.13 & 56.88 & $-4.35$ & --- \\
& Sonnet 4.6 (Direct Bbox) & 35.18 & 42.31 & $-33.64$ & 7.54 \\
& Qwen3-VL-8B (Direct) & 54.26 & 54.26 & $-54.26$ & 3.38 \\
& Qwen3-VL-32B (Direct) & 48.32 & 47.26 & $-47.26$ & 9.98 \\
\midrule
\textit{Specialist} & Grounding DINO (thr=0.10) & 52.76 & 77.61 & $+27.44$ & 32.19 \\
& CountGD (vanilla) & 43.01 & 126.82 & $+32.05$ & 36.13 \\
& CountGD (Crop+NMS) & 48.84 & 156.13 & $+42.52$ & 36.59 \\
\midrule
\textit{Heuristics} & SAHI (DINO) & 46.57 & 50.42 & $-44.65$ & 15.91 \\
& Hand-designed routing (DINO + slice) & 47.13 & 54.75 & $-45.15$ & 19.15 \\
\midrule
\textit{Agentic} & ReAct (Sonnet 4.6 + DINO) & 46.31 & 52.60 & $-46.03$ & 19.66 \\
& CodeAct (Sonnet 4.6 + tools) & 39.38 & 49.66 & $-32.40$ & 29.21 \\
& Reflexion (Sonnet 4.6 + visual critic) & 44.28 & 56.54 & $-25.88$ & 24.91 \\
\midrule
\textit{\methodname{} (ours)} & Analyzer OFF & 32.83 & 47.86 & $-16.39$ & 36.75 \\
& Analyzer ON & \textbf{23.36} & \textbf{37.77} & {\boldmath$-4.02$} & \textbf{40.48} \\
\bottomrule
\end{tabular}}
\begin{flushleft}
\footnotesize{\methodname{} (Analyzer ON) wins MAE, RMSE, Bias, and mIoU. The per-density-tier breakdown is in Table~\ref{tab:sota_per_density} (Appendix).}
\end{flushleft}
\end{table*}

\section{Evaluation Setup}
\label{sec:benchmark}

We evaluate \methodname{} on two visual reasoning tasks whose static-tool failure modes are well-characterized but orthogonal: dense object counting stresses calibration of detection thresholds and non-maximum suppression under occlusion, while grounded segmentation stresses spatial grounding and pixel-precise boundary recovery on out-of-distribution categories.  Both tasks derive ground truth from existing annotations (LVIS bounding boxes; PlantSeg pixel masks), enabling rigorous evaluation without subjective labeling (Figure~\ref{fig:eval_setup}, Appendix~\ref{sec:appendix_benchmark_fig}).

\subsection{Counting: LVIS-Count}
\label{sec:pillar_counting}

The counting task uses \textbf{LVIS-Count}, a 180-image subset sampled from LVIS~\citep{gupta2019lvis} spanning 30 categories across five semantic domains (60 train / 20 val / 100 zero-shot test). Test images fall into two density regimes: \textbf{Normal} ($10 \le N \le 50$), where threshold and NMS calibration dominate, and \textbf{Extreme} ($51 \le N \le 100$), where heavy occlusion forces algorithmic strategy shifts. We exclude the Sparse regime ($N<10$) because zero-shot detectors already saturate it, giving no discriminative signal between methods.

\subsection{Segmentation: PlantSeg-OOD}
\label{sec:pillar_segmentation}

The segmentation task uses \textbf{PlantSeg-OOD}, a 180-image species-disjoint subset of the public PlantSeg benchmark~\citep{wei2024plantseg}, split 60/20/100 train/val/test across 24 plant species, with the 10 test species held out from the 14 train/val species, so the 100-image test set evaluates \methodname{} on 21 disease classes from 10 plant species never seen during search. Lesions are stratified by mask-area ratio into \textbf{Small} (area $<10\%$, discovery is the bottleneck) and \textbf{Moderate} ($10\%$--$30\%$, boundary precision is the bottleneck) tiers; the original \textit{large} tier ($\ge 30\%$) is excluded so reported scores cannot be inflated by trivially-large foreground area. The 100-image test split contains 56 \textit{small} and 44 \textit{moderate} lesions.

\subsection{Evaluation Metrics}
\label{sec:metrics}

For counting we report \textbf{mIoU}, the mean Intersection-over-Union between predicted and ground-truth bounding boxes after a Hungarian-style one-to-one match constrained to symmetric center-in-box hits, and \textbf{MAE}, the mean absolute error in object count per image.  We additionally report counting Bias (signed mean error), RMSE (root-mean-square count error), and per-density-tier breakdowns.

For segmentation we report \textbf{Dice}$^{\text{mask}}$ (pixel-level Dice coefficient between predicted and ground-truth masks; our primary segmentation metric) and \textbf{mIoU}$^{\text{mask}}$ (pixel-level IoU averaged across instances).  Per-tier breakdowns (\textit{small} / \textit{moderate}) expose where each method fails.

\section{Experiments}

\begin{table*}[t]
\centering
\small
\caption{\textbf{PlantSeg-OOD zero-shot results.} Dice$^{\text{mask}}$ (\%) on the 100-image OOD test set; species and diseases are entirely held out from train/val. ``All'' = mean over the test set; ``Mod.'' / ``Small'' = stratified means on the two difficulty tiers. \methodname{}~Base is the configuration without analyzer-loop diagnostics; \methodname{} adds the Analyzer step. Bold = best in column.}
\label{tab:plantseg_v2}
\begin{tabular}{llcccc}
\toprule
\textbf{Method} & \textbf{LLM} & Dice All$\uparrow$ & Dice Mod.$\uparrow$ & Dice Small$\uparrow$ & mIoU All\%$\uparrow$ \\
\midrule
\multicolumn{6}{l}{\textit{Detector-only baselines}} \\
SAM 2 (segment-everything) & --- & 20.05 & 30.76 & 11.64 & 3.71 \\
SAM 2 + CLIP (vanilla zero-shot) & --- & 23.11 & 32.04 & 16.09 & 5.75 \\
Grounded-SAM 2 (zero-shot) & --- & 36.56 & \textbf{51.24} & 25.03 & 9.75 \\
\midrule
\multicolumn{6}{l}{\textit{VLM zero-shot}} \\
VLM-direct & GPT-4o-mini & 17.30 & 21.53 & 13.98 & 4.61 \\
VLM-direct & Sonnet 4.6 & 30.45 & 42.18 & 21.24 & 9.22 \\
\midrule
\multicolumn{6}{l}{\textit{Iterative / agentic}} \\
Reflexion ($K{=}3$) & Sonnet 4.6 & 31.62 & 38.14 & 26.50 & 10.57 \\
VisProg & Sonnet 4.6 & 36.33 & 45.79 & 28.90 & \textbf{14.27} \\
\midrule
\textbf{\methodname{} Base (Ours)} & Sonnet 4.6 & 36.79 & 45.30 & 30.10 & 12.84 \\
\textbf{\methodname{} (Ours)} & Sonnet 4.6 & \textbf{39.38} & 50.00 & \textbf{31.03} & 13.48 \\
\bottomrule
\end{tabular}
\end{table*}

\subsection{Experimental Setup}
We evaluate on the two tasks described in \S\ref{sec:benchmark} using the metrics defined in \S\ref{sec:metrics}. The 80-image LVIS-Count training split is used for algorithm search; all results below are on the held-out 100-image zero-shot test set. We compare against four baseline families: (i) direct end-to-end VLM prompting (Sonnet 4.6 Count-Only and Direct-Bbox; Qwen3-VL-8B and 32B Direct), (ii) specialist counting/detection models (Grounding DINO~\citep{liu2023grounding} at varying thresholds; CountGD~\citep{amini-naieni2024countgd} vanilla and Crop+NMS), (iii) heuristic pipelines (SAHI-style slice-and-detect~\citep{akyon2022slicing}), and (iv) agentic baselines (ReAct~\citep{yao2023react}, CodeAct~\citep{wang2024executable}, Reflexion~\citep{shinn2023reflexion} with visual critique). \methodname{}'s search budget defaults to $N{=}15$ iterations (Appendix~\ref{sec:appendix_ablations}), with $N{=}25$ for ceiling numbers with the Analyzer module enabled. Because no random seed is fixed and program synthesis is stochastic, independent runs of the same configuration can synthesize different programs and therefore yield different numbers; Table~\ref{tab:sota_comparison} reports a single conservative run, while the appendix ablations report separate runs.

\subsection{Counting: LVIS-Count}
\label{sec:lvis_counting}

Table~\ref{tab:sota_comparison} reports aggregate zero-shot test-set metrics across four baseline families and two \methodname{} configurations; per-density stratification of the same 13 methods is in Table~\ref{tab:sota_per_density} (Appendix).

\textbf{\methodname{} dominates the aggregate metrics.} The full \methodname{} (Analyzer ON, $N{=}15$) wins MAE (23.36), RMSE (37.77), Bias ($-4.02$), and mIoU (40.48) on the aggregate test set, the four metrics that jointly measure counting accuracy, scale-aware error, signed bias, and box localization. \methodname{}'s improvement over the strongest specialist (CountGD vanilla) is $+4.4$ mIoU (40.48 vs.\ 36.13) at 54\% of its MAE (23.36 vs.\ 43.01); direct-Bbox VLM prompting (Sonnet 4.6 or Qwen3-VL) trails at 3--10\% mIoU, confirming end-to-end VLMs cannot ground at this scale. On the per-density breakdown, \methodname{} (Analyzer ON) wins Normal mIoU (44.24), Normal MAE (17.28), and Extreme MAE (29.44), while CountGD Crop+NMS edges Extreme mIoU (37.06 vs 36.73), motivating the Analyzer-budget ablation in \S\ref{sec:ablations}.

\textbf{The gains stem from program search, not the tool pool.} A natural question is whether \methodname{} outperforms baselines primarily due to the expressive tool pool rather than the search over orchestration programs. We isolate this factor with a \emph{validation-tuned dispatch} control that uses the identical tool pool and operators as \methodname{} but replaces the searched program with a hand-designed, density-based routing policy (Grounding DINO with NMS on low-density scenes, slice-and-detect on high-density scenes, alongside area and confidence cutoffs). To ensure a rigorous and competitive control, all thresholds in this policy are grid-searched on the same $20$-image validation split under the identical Dual-Rank objective that \methodname{} optimizes, and the resulting policy is evaluated on the held-out $100$-image test set.

If the tool pool alone accounted for the performance gains, this baseline would close most of the gap. However, it achieves an MAE of $47.13$ and mIoU of $19.15$, performing on par with the heuristic SAHI baseline ($46.57$ / $15.91$) and falling far short of \methodname{} ($23.36$ / $40.48$). The core difference lies in the optimization target: while the dispatch baseline fixes the program structure and only tunes numerical thresholds within a static template, \methodname{} searches the control flow itself. The winning program synthesized non-trivial control structures beyond the expressivity of fixed templates, including a two-threshold consistency probe that separates under- from over-detection and a fusion mechanism combining slice-level detections with the global pass.

\textbf{Component contribution.} Within \methodname{}, the Analyzer module contributes the largest gain: enabling the Analyzer adds $+3.7$ aggregate mIoU and cuts MAE by 9.5 points ($32.83 \to 23.36$).

\subsection{Cross-Domain Generalization: PlantSeg-OOD}
\label{sec:plantseg}
To probe transfer beyond counting, we benchmark on PlantSeg-OOD (see \S\ref{sec:pillar_segmentation} for dataset details) against detector-only baselines (SAM~2~\citep{ravi2024sam}, SAM~2{+}CLIP~\citep{radford2021learning}, and Grounded-SAM~2), agentic baselines (VisProg, Reflexion), and direct VLM prompting. Table~\ref{tab:plantseg_v2} reports zero-shot Dice$^{\text{mask}}$ and mIoU$^{\text{mask}}$ on the 100-image OOD test split (56 \emph{small} + 44 \emph{moderate} lesions).

\noindent\textbf{Main segmentation result.} \methodname{} attains the best Dice$^{\text{mask}}$ on the \emph{All} set ($39.38$ vs.\ $36.56$ for Grounded-SAM~2) and on the \emph{Small} tier ($31.03$ vs.\ $28.90$ VisProg, $25.03$ Grounded-SAM~2), where the Producer--Analyzer advantage is largest: static detectors frequently miss lesions below 10\% of image area.  Adding the Analyzer loop to \methodname{}~Base lifts Dice~All by $+2.59$ and Dice~Small by $+0.93$.

\subsection{Efficiency}
\label{sec:efficiency}

A complete \methodname{} search incurs a one-off, offline cost of approximately $45$ LLM calls per task. Once searched, the resulting executable program composes vision tools directly without invoking any VLM, thus requiring \emph{zero} LLM calls per image at test time. In contrast, agentic baselines incur LLM costs on every query (at most $6$ calls per image for ReAct, $5$ for Reflexion, and $1$ for CodeAct). Across the $100$-image test set, \methodname{} consumes $45$ total calls, whereas ReAct, Reflexion, and CodeAct require up to $600$, $500$, and $100$ calls, respectively.

Because \methodname{} pays this cost once while baselines pay per image, the cost comparison exhibits a crossover rather than a fixed ratio. Specifically, \methodname{} achieves a lower cumulative cost than ReAct after $8$ images, Reflexion after $9$ images, and CodeAct after $45$ images; our $100$-image evaluation is therefore already past every crossover, and the advantage grows with deployment size. Figure~\ref{fig:cost_crossover} in Appendix~\ref{sec:appendix_efficiency} plots the cumulative call counts across the benchmark.

\subsection{Ablation Summary}
\label{sec:ablations}
We report four verified ablations in Appendix~\ref{sec:appendix_extended_results}: (i) search-iteration budget $N \in \{5, 10, 15, 20\}$; (ii) Analyzer Diagnostic toggle; (iii) per-iteration proposal count $K \in \{1, 3\}$; (iv) DINO confidence-threshold sensitivity. The Analyzer ablation is the key finding: at fixed budget $N{=}10$, the Analyzer cuts MAE by 36\% ($33.51 \to 21.38$) and drives Bias from $+11.6$ to near-zero ($+0.6$); raising the budget to $N{=}25$ attains the best mIoU in the experiment matrix ($41.66$). Appendix~\ref{sec:appendix_extended_results} also reports the per-density counting breakdown across all 14 methods (Table~\ref{tab:sota_per_density}), a density-tier mIoU trajectory under Producer--Analyzer search, and a qualitative case-study walkthrough of a single search run.

\section{Conclusion}
\label{sec:conclusion}

We presented \methodname{}, which treats vision tools as instruments needing scene-specific
calibration rather than reliable oracles, and searches for the orchestration instead of
assuming it. A Producer loop proposes candidate programs over a constrained tool library, an
Analyzer loop proposes diagnostic programs that expose how they fail, and a shared Knowledge
Base carries both across iterations. On dense counting (LVIS-Count) and out-of-distribution
segmentation (PlantSeg-OOD), the searched programs improve over direct VLM prompting,
specialist detectors and program-synthesis baselines, and the deployed artifact runs without
a VLM.

\section*{Limitations}

\textbf{Scope of evaluation.} Our experiments cover two visual-reasoning tasks: dense object counting on LVIS-Count (180 images, 30 categories, English category labels grouped into 5 semantic domains) and zero-shot grounded segmentation on PlantSeg-OOD (180 images across 24 plant species; the 10 test species are held out from train/val). Generalization to other visual domains (medical imaging, satellite, document understanding), to languages other than English, and to compositional queries that mix counting with relational reasoning is not evaluated; we expect degradation under each of these shifts.

\textbf{Compute and reproducibility.} \methodname{} performs offline search with no parameter updates, so reproduction requires no GPU training infrastructure. A full $N{=}15$ search run under the Analyzer-ON configuration makes approximately $45$ calls to a \emph{single} LLM (Claude Sonnet~4.6, shared across the Producer, Analyzer, and Knowledge-Base update rather than using an ensemble of models), alongside subprocess evaluations against the VFM toolbox for each task. We clarify this accounting to avoid double-counting: a single Producer call yields all $K{=}3$ candidate programs in one response, meaning costs scale with the number of \emph{LLM calls} rather than the number of \emph{generated programs} (which would otherwise suggest a loose upper bound of ${\approx}90$). Reduced-budget configurations ($N{=}5$, Analyzer OFF) achieve most of the performance gains at proportionally lower search cost (Appendix~\ref{sec:appendix_extended_results}).

\textbf{Applicability.} \methodname{} provides little benefit on simple tasks where a single calibrated tool already operates near saturation (such as sparse scenes with few isolated instances). In such settings, static pipelines remain preferred as they incur zero search overhead without sacrificing accuracy.

\textbf{Tool-use safety.} \methodname{} delegates execution to external VFM tools (Grounding DINO, SAM, NMS) and to an LLM API. The benchmark does not include adversarial tool inputs (e.g., images crafted to cause specific NMS-threshold failures); production deployment would require additional sandboxing of the subprocess evaluator, rate-limiting on the LLM/API budget, and audit-logging of the deployed program output.

\textbf{Reasoning ceiling.} Performance is bounded by the intersection of three capability ceilings: the VLM's diagnostic fidelity during Analyzer steps, the LLM's program-synthesis capability during Producer steps, and the VFM toolbox's perceptual expressiveness at deployment. When a failure mode falls outside this intersection the Knowledge Base records it as an \textsc{Unverified} hypothesis, but the search cannot self-correct beyond its tool repertoire; disentangling these three ceilings empirically is left to future work. Moreover, the tool library is fixed: \methodname{} composes and calibrates existing primitives but cannot invent perceptual operators beyond those provided.

\section*{Acknowledgments}
We thank the anonymous reviewers for their valuable comments and suggestions.
This research was supported by the Australian Government through the Australian Research Council's Discovery Projects funding scheme (project DP220101784).

\bibliography{references}

\clearpage
\appendix

\section{Evaluation Setup Figure}
\label{sec:appendix_benchmark_fig}

\begin{figure*}[t]
\centering
\includegraphics[width=\textwidth]{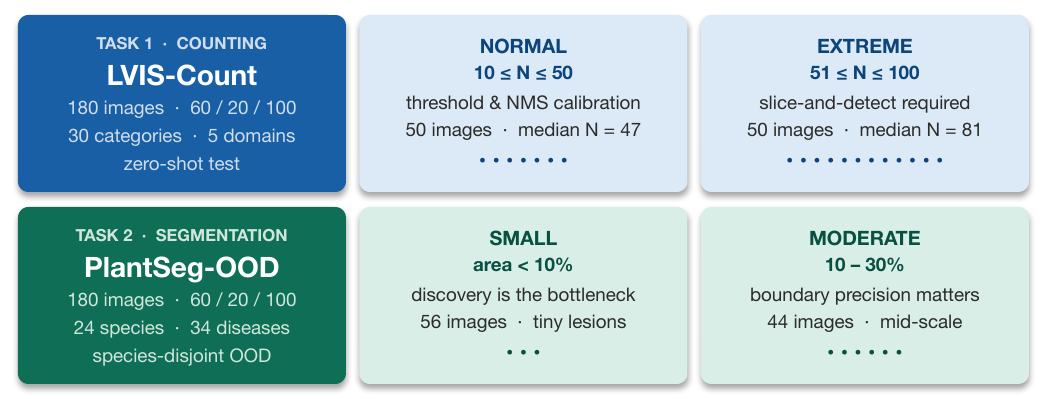}
\caption{\textbf{Two-task evaluation setup.} \methodname{} is evaluated on dense object counting (top, LVIS-Count: 180 images sampled from LVIS, 30 categories, two density tiers --- Normal $10\!\le\!N\!\le\!50$ and Extreme $51\!\le\!N\!\le\!100$) and on grounded segmentation (bottom, PlantSeg-OOD: 100-image OOD test sampled from PlantSeg, three-way species-disjoint splits, two lesion-area tiers --- Small $<\!10\%$ and Moderate $10$--$30\%$). Both tasks expose distinct failure modes of static tool invocation.}
\label{fig:eval_setup}
\end{figure*}

For reference, Figure~\ref{fig:eval_setup} gives a visual overview of the two evaluation tasks; tier definitions and split details are in \S\ref{sec:benchmark}.

\section{Extended Related Work}
\label{sec:appendix_related}

\paragraph{Adaptive Perception and Tool Calibration.} Fixed vision models require calibration to generalize across diverse domains—a challenge addressed by test-time adaptation methods that update model weights to mitigate distribution shift~\citep{wang2021tent}. For tool-augmented agents, however, the bottleneck is not weight adaptation but \textit{parameter-level sensitivity}: atomic tools such as object detectors depend critically on hyper-parameters (confidence thresholds, NMS values) that vary per image. Counting-oriented methods~\citep{ranjan2021learning} further underscore this difficulty by revealing wide performance variance across density regimes. While frameworks like HuggingGPT~\citep{shen2023hugginggpt} explore dynamic tool selection, they do not perform per-instance parameter tuning. \methodname{} fills this niche by modeling vision tools as \textit{dynamic instruments} whose configurations the searched program selects per image, without requiring any parameter updates to the underlying models or any model call at inference.

\paragraph{Prompt-Tuning and Differentiable Text Optimization.} TextGrad~\citep{yuksekgonul2024textgrad} introduces automatic differentiation over text to optimize compound AI systems, while system-level prompt-optimization approaches such as MetaPrompting~\citep{hou2023metaprompting} and MetaSPO~\citep{choi2025system} explore meta-learning frameworks for robust prompt design. These methods optimize prompts rather than executable code; \methodname{} differs in that it searches over executable Python programs and uses deterministic interpreter feedback as the correction signal (\S\ref{sec:related}).

\section{Benchmark Statistics}
\label{sec:appendix_stats}

The two datasets total 360 images; the 100-image LVIS-Count test split is balanced 50/50 between the Normal ($10\!\le\!N\!\le\!50$) and Extreme ($51\!\le\!N\!\le\!100$) tiers, and the 100-image PlantSeg-OOD test split contains 56 \emph{small} and 44 \emph{moderate} lesions. Full per-task statistics are in \S\ref{sec:benchmark}.

\section{Implementation Mechanisms Not Detailed in the Main Method}
\label{sec:appendix_extra_mechanisms}
For reviewers and reproducibility, we document four engineering mechanisms present in the released code but elided from the main Method. Both Producer- and Analyzer-proposed programs run in isolated Python subprocesses; the evaluator returns metrics as JSON over stdout, and a crashed subprocess records a score of $-\infty$ with a traceback into its snapshot. A hard cap of five VLM calls per analyzer execution keeps the Analyzer from degenerating into an expensive Visual-QA wrapper. The full \texttt{VisionSearchState} is serialised to \texttt{records.json} after each iteration, so an interrupted run resumes from the last completed step. Finally, an AST-level lint pass rejects out-of-library imports before execution, preventing the LLM from hallucinating new tools and keeping deployed programs portable.

\section{Discussion}
\label{sec:discussion}
\methodname{}'s advantage comes from treating vision tools as scene-sensitive dynamic instruments: the Producer--Analyzer search discovers per-scene parameter combinations (DINO threshold $0.10$--$0.15$, tighter NMS IoU $\approx 0.3$, morphological dilation) through hypothesis-driven trial-and-error that one-shot program synthesis cannot. CountGD edges Extreme-tier mIoU by 0.3pp because it is a purpose-built dense-scene counter, but pays with $+32.05$ Bias and aggregate MAE nearly double \methodname{}'s, as the two systems reward orthogonal capabilities. The default $N{=}15$ budget saturates the Analyzer-OFF configuration; enabling the Analyzer extends gains up to $N{=}25$, beyond which compute yields diminishing returns.

Listing~\ref{lst:analyzer_excerpt} (Appendix~\ref{sec:appendix_code}) gives a representative analyzer program; to show what such diagnostics buy, we trace the PlantSeg-OOD run behind Table~\ref{tab:plantseg_v2} (its full trajectory is in Appendix~\ref{sec:appendix_code}). The Analyzer's observations isolate a sequence of distinct failure modes: small lesions go undetected at the seed's high detection threshold, SAM mask boundaries come back ragged, and a bare disease query mislocalises on unfamiliar plant species. The search converts each observation into a targeted, auditable revision: lowering the detection threshold to $0.15$ with an area filter, adding \texttt{clean\_mask} boundary post-processing, grounding the query as ``\emph{disease} on \emph{plant}'', and loosening NMS to preserve co-located lesions, which the Producer composes into its best program, raising this run's training-split Dice from $32.90$ to $40.00$. Aggregated over the held-out test set, the same diagnose$\to$revise loop yields the $+2.59$ Dice-All / $+0.93$ Dice-Small gain of \methodname{} over \methodname{}~Base in Table~\ref{tab:plantseg_v2}, turning an opaque score gap into auditable code edits.

\section{Efficiency Crossover}
\label{sec:appendix_efficiency}

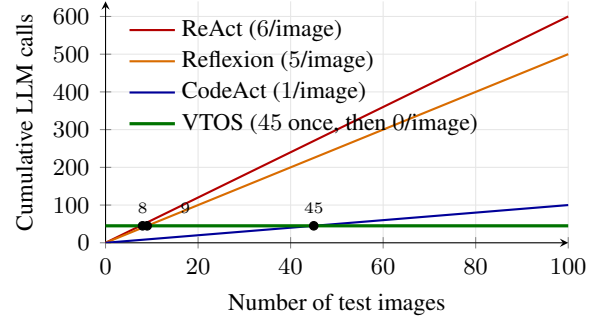
\begin{figure}[t]
\centering
\begin{tikzpicture}
\begin{axis}[
  width=\columnwidth, height=0.62\columnwidth,
  xlabel={Number of test images},
  ylabel={Cumulative LLM calls},
  xmin=0, xmax=100, ymin=0, ymax=640,
  xtick={0,20,40,60,80,100},
  ytick={0,100,200,300,400,500,600},
  legend pos=north west,
  legend cell align=left,
  legend style={font=\footnotesize, draw=none, fill=none},
  label style={font=\small}, tick label style={font=\footnotesize},
  grid=major, grid style={gray!20},
  axis lines=left,
]
\addplot[thick, red!70!black]      coordinates {(0,0) (100,600)}; \addlegendentry{ReAct ($6$/image)}
\addplot[thick, orange!85!black]   coordinates {(0,0) (100,500)}; \addlegendentry{Reflexion ($5$/image)}
\addplot[thick, blue!60!black]     coordinates {(0,0) (100,100)}; \addlegendentry{CodeAct ($1$/image)}
\addplot[very thick, green!45!black] coordinates {(0,45) (100,45)}; \addlegendentry{\methodname{} ($45$ once, then $0$/image)}

\addplot[only marks, mark=*, mark size=1.6pt, black] coordinates {(8,45) (9,45) (45,45)};
\node[font=\tiny, anchor=south]      at (axis cs:8,52)  {$8$};
\node[font=\tiny, anchor=south west] at (axis cs:14,52) {$9$};
\node[font=\tiny, anchor=south]      at (axis cs:45,52) {$45$};
\end{axis}
\end{tikzpicture}
\caption{\textbf{Cumulative LLM cost over the $100$-image zero-shot test set.}
\methodname{} incurs a one-off offline search cost (${\approx}45$ calls) and deploys a tool-only program with zero test-time LLM calls, resulting in a flat cost trajectory. In contrast, agentic baselines pay per image, growing linearly. Dots mark the cost crossover points ($8$ images for ReAct, $9$ for Reflexion, and $45$ for CodeAct), beyond which \methodname{} becomes strictly more cost-effective.
}
\label{fig:cost_crossover}
\end{figure}

We report cost as the number of LLM calls, which reflects the intrinsic algorithmic requirement of each method independent of external API pricing or rate limits. \methodname{} incurs approximately $45$ calls as a one-off offline search cost; the resulting artifact is an executable program over vision tools that invokes no VLM at inference, yielding a per-image marginal cost of zero. Baseline per-image costs follow the configurations in Table~\ref{tab:sota_comparison}: ReAct is capped at $6$ tool-use turns per image, Reflexion uses $5$ (one initial attempt plus four reflection rounds), and CodeAct issues a single generation per image. \methodname{} therefore overtakes ReAct after $8$ images, Reflexion after $9$, and CodeAct after $45$.

\section{Extended Empirical Results}
\label{sec:appendix_extended_results}

\subsection{Qualitative Detection Examples}
\label{sec:appendix_detection_examples}
\begin{figure*}[t]
\centering
\includegraphics[width=\textwidth]{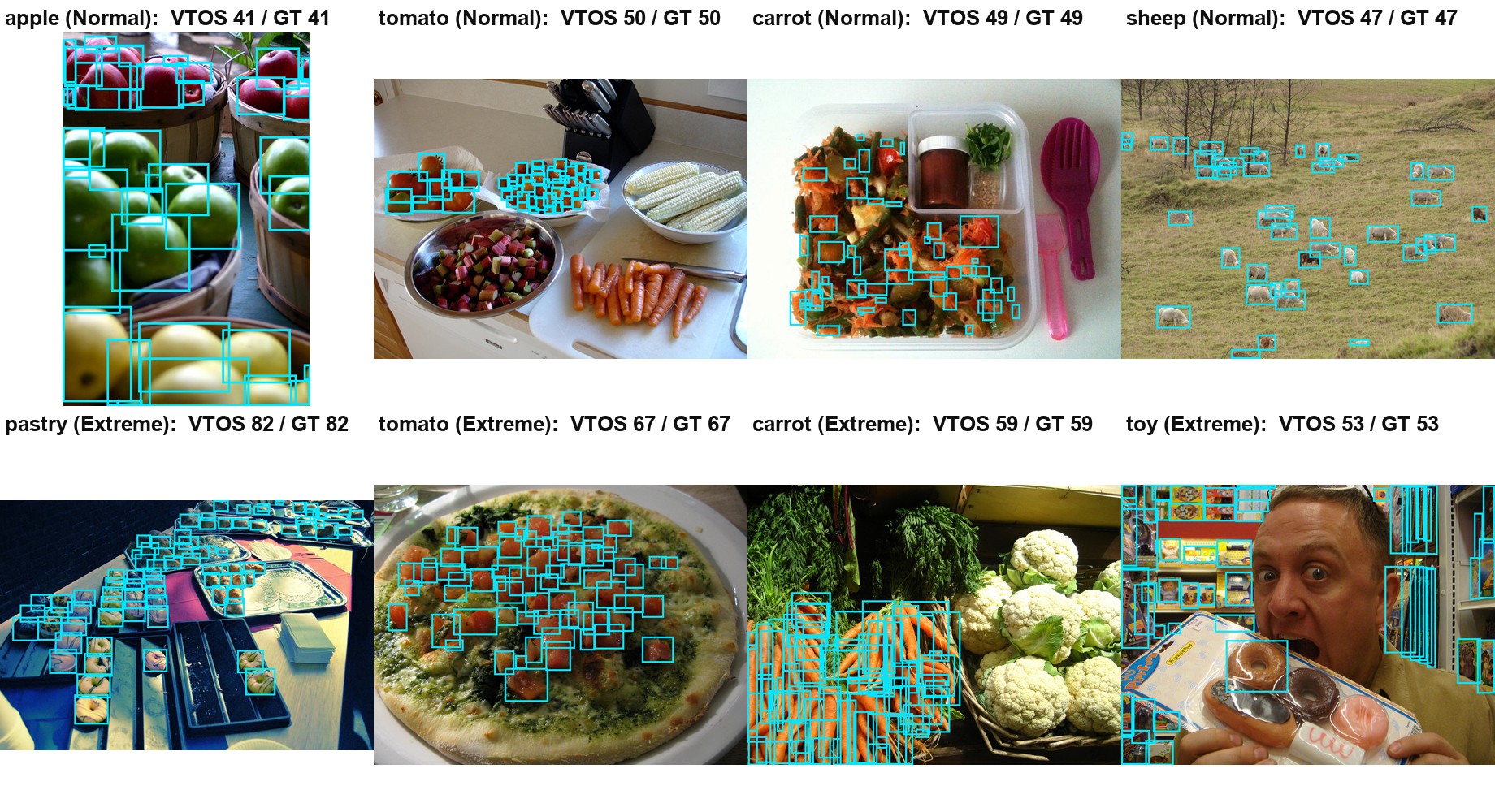}
\caption{\textbf{\methodname{} detection examples on LVIS-Count (zero-shot test set).} Top row: Normal tier ($10\,{\le}\,N\,{\le}\,50$); bottom row: Extreme tier ($51\,{\le}\,N\,{\le}\,100$). Cyan boxes mark ground-truth instances; each panel reports the \methodname{} program's predicted count versus ground truth. These representative cases show the searched counting program recovering the correct count across object categories and density regimes.}
\label{fig:detection_examples}
\end{figure*}

Figure~\ref{fig:detection_examples} shows representative \methodname{} counting results on the zero-shot LVIS-Count test set, spanning eight object categories and both density tiers. Ground-truth instance boxes are overlaid in cyan; for each panel the count returned by \methodname{}'s searched program matches the ground-truth count, illustrating that the discovered orchestration generalizes across diverse categories (fruit, vegetables, animals, baked goods, toys) and crowding levels.

\begin{table*}[ht]
\centering
\small
\caption{\textbf{Per-Density Breakdown on LVIS-Count.} Same 13 methods as Table~\ref{tab:sota_comparison}, stratified into Normal ($10 \le N \le 50$) and Extreme ($51 \le N \le 100$) density tiers. Bias is signed count error; zero is best. Best-per-column-within-tier in \textbf{bold}.}
\label{tab:sota_per_density}
\resizebox{\textwidth}{!}{%
\begin{tabular}{@{}llcccccc@{}}
\toprule
& & \multicolumn{3}{c}{\textbf{Normal} ($10 \le N \le 50$)} & \multicolumn{3}{c}{\textbf{Extreme} ($51 \le N \le 100$)} \\
\cmidrule(lr){3-5} \cmidrule(lr){6-8}
\textbf{Family} & \textbf{Method} & MAE$\downarrow$ & Bias ($\downarrow |\cdot|$) & mIoU\%$\uparrow$ & MAE$\downarrow$ & Bias ($\downarrow |\cdot|$) & mIoU\%$\uparrow$ \\
\midrule
\textit{End-to-End VLM} & Sonnet 4.6 (Count Only) & 24.56 & $-3.40$ & --- & 45.70 & {\boldmath$-5.30$} & --- \\
& Sonnet 4.6 (Direct Bbox) & 22.56 & $-19.80$ & 8.44 & 47.80 & $-47.48$ & 6.65 \\
& Qwen3-VL-8B (Direct) & 35.58 & $-35.58$ & 5.24 & 72.94 & $-72.94$ & 1.52 \\
& Qwen3-VL-32B (Direct) & 31.78 & $-47.26$ & 9.98 & 64.80 & $-64.80$ & 6.20 \\
\midrule
\textit{Specialist} & Grounding DINO (thr=0.10) & 54.30 & $+40.90$ & 33.08 & 51.22 & $+13.98$ & 31.30 \\
& CountGD (vanilla) & 45.32 & $+36.80$ & 36.07 & 40.70 & $+27.30$ & 36.19 \\
& CountGD (Crop+NMS) & 51.28 & $+46.72$ & 36.12 & 46.40 & $+38.32$ & \textbf{37.06} \\
\midrule
\textit{Heuristics} & SAHI (DINO) & 29.94 & $-26.66$ & 21.25 & 63.20 & $-62.64$ & 10.57 \\
\midrule
\textit{Agentic} & ReAct (Sonnet 4.6 + DINO) & 30.20 & $-29.64$ & 25.47 & 62.42 & $-62.42$ & 13.85 \\
& CodeAct (Sonnet 4.6 + tools) & 26.00 & $-14.80$ & 35.01 & 52.76 & $-50.00$ & 23.41 \\
& Reflexion (Sonnet 4.6 + visual critic) & 34.18 & $-10.98$ & 29.23 & 54.38 & $-40.78$ & 20.58 \\
\midrule
\textit{\methodname{} (ours)} & Analyzer OFF & 24.18 & {\boldmath$-1.18$} & 41.62 & 41.48 & $-31.60$ & 31.87 \\
& Analyzer ON & \textbf{17.28} & $+9.44$ & \textbf{44.24} & \textbf{29.44} & $-17.48$ & 36.73 \\
\bottomrule
\end{tabular}}
\begin{flushleft}
\footnotesize{\methodname{} (Analyzer ON) wins Normal MAE, Normal mIoU, and Extreme MAE; \methodname{} (Analyzer OFF) wins Normal Bias. CountGD (Crop+NMS) edges Extreme mIoU. Bold marks the best entry per column within its tier.}
\end{flushleft}
\end{table*}

\subsection{Ablation Studies}
\label{sec:appendix_ablations}

We report four verified ablations: (i) scaling the per-task search iteration budget; (ii) toggling the Analyzer diagnostic module; (iii) varying the per-iteration proposal count $K$; and (iv) DINO confidence-threshold sensitivity.

\begin{table*}[ht]
\centering
\small
\caption{\textbf{Ablation 1 --- Scaling Search Iterations.} \methodname{} (Analyzer OFF) at $N \in \{5, 10, 15, 20\}$. Aggregate over the full LVIS-Count test set.}
\label{tab:ablation_iterations}
\begin{tabular}{lcccc}
\toprule
Config & mIoU$\uparrow$ & MAE$\downarrow$ & RMSE$\downarrow$ & Bias ($\downarrow |\cdot|$) \\
\midrule
$N{=}5$  & \textbf{38.09} & 30.93 & \textbf{51.37} & $-3.71$ \\
$N{=}10$ & 35.05 & 33.51 & 53.63 & $+11.57$ \\
$N{=}15$ & 36.74 & \textbf{30.34} & 53.95 & $+7.20$ \\
$N{=}20$ & 36.37 & 32.03 & 54.68 & {\boldmath$+2.70$} \\
\bottomrule
\end{tabular}
\end{table*}

\begin{table*}[ht]
\centering
\small
\caption{\textbf{Ablation 2 --- Analyzer Diagnostic Module.} Two control regimes: same-budget ($N{=}10$) and ceiling-budget (Analyzer-OFF best at $N{=}15$; Analyzer-ON ceiling at $N{=}25$).}
\label{tab:ablation_analyzer}
\begin{tabular}{llccc}
\toprule
Regime & Config & mIoU$\uparrow$ & MAE$\downarrow$ & Bias ($\downarrow |\cdot|$) \\
\midrule
Same budget ($N{=}10$) & Analyzer OFF & 35.05 & 33.51 & +11.57 \\
& Analyzer ON  & 39.82 & 21.38 & {\boldmath$+0.56$} \\
\midrule
Ceiling & Analyzer OFF ($N{=}15$) & 36.74 & 30.34 & $+7.20$ \\
& Analyzer ON ($N{=}25$) & \textbf{41.66} & \textbf{19.63} & $+4.15$ \\
\bottomrule
\end{tabular}
\end{table*}

The Analyzer ablation is the key finding: at fixed budget $N{=}10$, the Analyzer cuts MAE by 36\% (33.51 $\to$ 21.38) and drives Bias from $+11.6$ to near-zero ($+0.6$). It also raises the achievable performance ceiling: Analyzer-OFF saturates at $N{=}15$ (additional iterations degrade mIoU), but Analyzer-ON continues to improve up to $N{=}25$, attaining $41.66$ mIoU, the best number in the entire experiment matrix.

\begin{table}[ht]
\centering
\small
\caption{\textbf{Ablation 3 --- Proposal Scaling $K$.} Effect of per-iteration proposal count. Analyzer-OFF base, $N{=}10$.}
\label{tab:ablation_K}
\begin{tabular}{lcc}
\toprule
Config & mIoU$\uparrow$ & MAE$\downarrow$ \\
\midrule
$K{=}1$ & 34.53 & 35.83 \\
$K{=}3$ (default) & \textbf{35.05} & \textbf{33.51} \\
\bottomrule
\end{tabular}
\end{table}

The $K{=}1$ vs $K{=}3$ comparison shows the cost-quality trade-off of parallel proposal generation: $K{=}3$ adds a small mIoU lift ($+0.5$) and reduces MAE by $\sim 6\%$ at $3{\times}$ the LLM cost per iteration. We default to $K{=}3$ in headline numbers.

\begin{table}[ht]
\centering
\small
\caption{\textbf{Ablation 4 --- DINO Confidence-Threshold Sensitivity.} mIoU/MAE on LVIS-Count as the DINO open-vocabulary detection threshold sweeps from 0.10 to 0.35.}
\label{tab:ablation_dino_threshold}
\begin{tabular}{lcc}
\toprule
DINO threshold & mIoU$\uparrow$ & MAE$\downarrow$ \\
\midrule
0.35 & 14.13 & 52.12 \\
0.30 & 18.37 & 48.21 \\
0.20 & 29.01 & \textbf{39.34} \\
0.10 & \textbf{32.19} & 52.76 \\
\bottomrule
\end{tabular}
\end{table}

The threshold sweep underscores the brittleness motivating \methodname{}: mIoU rises monotonically from $14.13$ at the COCO-default threshold of $0.35$ to $32.19$ at threshold $0.10$, but MAE bottoms out at threshold $0.20$ (39.34) and rebounds at $0.10$ (52.76) because the lowest threshold floods the scene with false positives. No single threshold is optimal on both metrics simultaneously, a per-scene calibration target that \methodname{}'s search is designed to discover.

\subsection{Evolutionary Trajectory Analysis}

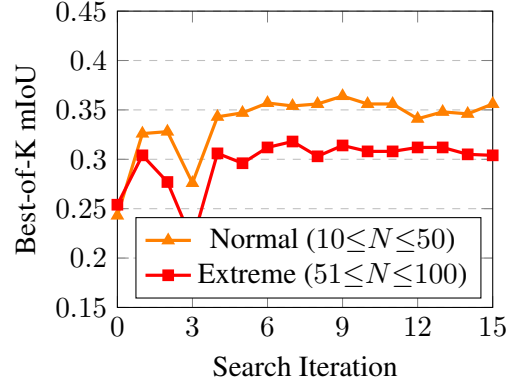
\begin{figure}[ht]
    \centering
    \begin{tikzpicture}
        \begin{axis}[
            title={\textbf{Self-Correction Trajectory (LVIS-Count)}},
            xlabel={Search Iteration},
            ylabel={Best-of-K mIoU},
            xmin=0, xmax=15,
            ymin=0.15, ymax=0.45,
            xtick={0,3,6,9,12,15},
            ytick={0.15,0.20,0.25,0.30,0.35,0.40,0.45},
            legend pos=south east,
            ymajorgrids=true,
            grid style=dashed,
            width=0.85\columnwidth,
            height=5.5cm,
        ]

        \addplot[
            color=orange,
            mark=triangle*,
            mark size=1.8pt,
            line width=1.2pt,
        ]
        coordinates {
            (0,0.243)(1,0.326)(2,0.328)(3,0.276)(4,0.343)(5,0.347)(6,0.357)(7,0.354)(8,0.356)(9,0.364)(10,0.356)(11,0.356)(12,0.341)(13,0.348)(14,0.346)(15,0.356)
        };
        \addlegendentry{Normal ($10{\le}N{\le}50$)}

        \addplot[
            color=red,
            mark=square*,
            mark size=1.6pt,
            line width=1.2pt,
        ]
        coordinates {
            (0,0.254)(1,0.304)(2,0.277)(3,0.217)(4,0.306)(5,0.296)(6,0.312)(7,0.318)(8,0.303)(9,0.314)(10,0.308)(11,0.308)(12,0.312)(13,0.312)(14,0.305)(15,0.304)
        };
        \addlegendentry{Extreme ($51{\le}N{\le}100$)}

        \end{axis}
    \end{tikzpicture}
    \caption{\textbf{mIoU Trajectory over Search Iterations} on the LVIS-Count training split. Per-iteration best-of-$K$ ($K{=}3$) mIoU stratified by density tier (Normal: $10{\le}N{\le}50$; Extreme: $51{\le}N{\le}100$). The trajectory is non-monotonic --- both tiers show occasional regressions (e.g., Extreme dips at iter~3 from $0.30$ to $0.22$) as the Producer explores new code directions in response to Analyzer feedback. Best Normal mIoU ($0.364$) is reached at iter~9, best Extreme ($0.318$) at iter~7; subsequent iterations plateau with small oscillations characteristic of search saturation.}
    \label{fig:evolution_trajectory}
\end{figure}

Figure~\ref{fig:evolution_trajectory} visualizes the self-correction trajectory across the two density regimes on the 60-image training split. Both regimes show clear improvement from the baseline (Normal $0.243\!\to\!0.364$; Extreme $0.254\!\to\!0.318$) but the trajectory is \emph{not} monotonic: at iter~3 the Producer proposes solutions that regress on both tiers (Extreme drops from $0.30$ to $0.22$) before recovering at iter~4. Such regressions correspond to Knowledge-Base updates in which an existing hypothesis is re-statused from \textsc{Unverified} to \textsc{Refuted} and a new strategy is proposed (e.g., switching from threshold-based NMS to morphological post-processing). After iter~9 the running best plateaus with small oscillations characteristic of search saturation.

\section{Future Work and Code/Data Availability}
\label{sec:appendix_future_and_release}

\paragraph{Future Work.} We see three natural extensions: (i) extending the search to additional visual reasoning tasks such as fine-grained attribute analysis and geometric distortion correction; (ii) integrating lightweight tool fine-tuning as a complement to parameter-level calibration; and (iii) investigating whether VisionThoughts accumulation across hundreds of diverse visual tasks exhibits emergent compositional capabilities. The bounded scope of evaluation, compute profile, and capability ceilings are discussed in the Limitations section.

\paragraph{Code and Data Availability.} Source code is available at \url{https://github.com/jinchaogjc/VTOS} under the MIT license. The full implementation, the LVIS-Count and PlantSeg-OOD dataset splits, and complete search-run logs (including the VisionThoughts Knowledge Bases and ranked solution snapshots) will be released there before the conference.

\section{Knowledge Base Field Definitions}
\label{sec:appendix_kb_fields}

The Knowledge Base holds six fields, $\mathcal{K} = \langle \mathcal{I}, \mathcal{H}, \mathcal{A}, \mathcal{E}, p, e \rangle$, all updated by the LLM each iteration from the previous state and the new evidence. The \emph{insights} $\mathcal{I}$ are empirical facts observed during search (e.g., that threshold $0.10$ maximizes mIoU on dense scenes but inflates MAE), while the \emph{hypotheses} $\mathcal{H}$ are $\langle$statement, status, evidence$\rangle$ triples, with status in \{\textsc{Confirmed}, \textsc{Refuted}, \textsc{Partial}, \textsc{Unverified}\}, that are re-assessed each iteration. The \emph{analyzer-insights} $\mathcal{A}$ track which analyses yield actionable findings so as to steer later analyzer proposals, and the \emph{exploration-directions} $\mathcal{E}$ are $\langle$direction, status, performance$\rangle$ triples for strategies tried, planned, or abandoned. The remaining two fields summarize search state: the \emph{key-problems} $p$ name the current bottlenecks in free text, and the \emph{exploration-status} $e$ is a label (improving, stuck, or converging) that biases the next iteration toward exploration or exploitation. The Knowledge Base is serialised to JSON each iteration, so an interrupted run resumes from the last completed step.

\section{How Co-Search Grows a Program: Two Trajectories}
\label{sec:appendix_code}

Open-loop visual-programming systems (e.g., ViperGPT, VisProg) emit a single fixed pipeline in one generation pass. Every \methodname{} program, by contrast, is \emph{searched}: each revision is traceable to a specific analyzer diagnosis of the previous attempt. We walk through the two paper runs --- one per task. For each, a table gives the diagnosis-driven trajectory at a glance, and the listings below show the actual code growing step by step. Scores along a chain are training-split metrics (mIoU for counting, Dice for segmentation); the selected programs attain the held-out test numbers of Tables~\ref{tab:sota_comparison} and~\ref{tab:plantseg_v2}.

\subsection{Counting Trajectory}
The paper's Analyzer-ON counting run grows a single Grounding DINO call into an image-adaptive program; its selected program attains test mIoU $40.48$ (Table~\ref{tab:sota_comparison}). Table~\ref{tab:count_evo} traces the search at a glance; the listings then show the code growing.

\begin{table*}[ht]
\centering\small
\caption{\textbf{Counting co-search}: each analyzer diagnosis drives a traceable revision, and the training mIoU rises monotonically toward the deployed program.}
\label{tab:count_evo}
\begin{tabular}{@{}c l l c@{}}
\toprule
it & Analyzer diagnosis & Solution revision & mIoU (\%) \\
\midrule
0  & --- (seed) & single Grounding DINO @\,0.10 & 24.84 \\
1  & duplicate boxes inflate the count & two-threshold union, then NMS & 31.58 \\
2  & crowded cells over-merge under fixed NMS & density-adaptive NMS by cell density & 32.42 \\
3  & images with $<$40 detections under-detect & conditional \texttt{slice\_and\_detect} ($n<40$) & 33.54 \\
12 & a dense image still under-detected (conf-ratio $0.83$) & learned ratio$+$count slicing triggers & 35.90 \\
\bottomrule
\end{tabular}
\end{table*}

\noindent\textbf{Seed} ($24.84$):
\begin{lstlisting}[style=pythoncode]
bboxes = detect(box_threshold=0.10)
\end{lstlisting}
\noindent$\hookrightarrow$ \textbf{$+$ two-threshold union} ($31.58$):
\begin{lstlisting}[style=pythoncode]
bboxes = nms(detect(0.10) + detect(0.20), iou=0.35)
\end{lstlisting}
\noindent$\hookrightarrow$ \textbf{$+$ density-adaptive NMS} ($32.42$):
\begin{lstlisting}[style=pythoncode]
max_cell = densest_3x3_cell(bboxes)
iou = 0.25 if max_cell > 8 else 0.35 if max_cell > 4 else 0.45
bboxes = nms(bboxes, iou)
\end{lstlisting}
\noindent$\hookrightarrow$ \textbf{$+$ conditional slicing} ($33.54$):
\begin{lstlisting}[style=pythoncode]
if len(bboxes) < 40:                    # under-detected -> recover by slicing
    bboxes = nms(bboxes + slice_and_detect(grid=(2,2), 0.15), 0.30)
\end{lstlisting}
\noindent$\hookrightarrow$ \textbf{Final} (test mIoU \textbf{40.48}):
\begin{lstlisting}[style=pythoncode]
ratio = n_high / max(n_low, 1)
if n_low < 32 or (n_low < 55 and ratio > 0.70 and n_high < 35):
    bboxes = nms(bboxes_low + slice_and_detect(grid=(3,3), 0.12), 0.28)
elif n_low > 55: bboxes = nms(detect(0.14), 0.25)   # dense: tighter threshold + NMS
else:            bboxes = nms(bboxes_low, 0.30)
\end{lstlisting}

\subsection{Segmentation Trajectory}
The Analyzer-ON segmentation run instead evolves query phrasing, thresholds, and mask post-processing; its selected program attains test Dice $39.38$ (Table~\ref{tab:plantseg_v2}). The strategies it discovers (Table~\ref{tab:seg_evo}) are an entirely different family from the counting chain --- co-search adapts to each task's distinct failure modes.

\begin{table*}[ht]
\centering\small
\caption{\textbf{Segmentation co-search}: a different family of revisions --- query phrasing and mask refinement rather than detection routing.}
\label{tab:seg_evo}
\begin{tabular}{@{}c l l c@{}}
\toprule
it & Analyzer diagnosis & Solution revision & Dice (\%) \\
\midrule
0 & --- (seed) & \texttt{detect}@0.30 then SAM & 32.90 \\
1 & small lesions missed at high threshold & lower threshold to $0.15$ $+$ area filter & 38.10 \\
2 & SAM mask boundaries are ragged & \texttt{clean\_mask} boundary post-processing & 38.10 \\
3 & bare query mislocalises across plants & contextual query (\emph{disease} on \emph{plant}) & 39.20 \\
4 & tight NMS merges co-located lesions & loosen NMS to $0.55$ & 40.00 \\
\bottomrule
\end{tabular}
\end{table*}

\noindent\textbf{Seed} ($32.90$):
\begin{lstlisting}[style=pythoncode]
bboxes   = detect(target, threshold=0.30)
polygons = segment(bboxes)                     # SAM
\end{lstlisting}
\noindent$\hookrightarrow$ \textbf{$+$ low threshold $+$ area filter} ($38.10$):
\begin{lstlisting}[style=pythoncode]
bboxes = detect(target, threshold=0.15)        # recover small lesions
bboxes = nms(filter_by_area(bboxes, 0.0003, 0.80), 0.40)
\end{lstlisting}
\noindent$\hookrightarrow$ \textbf{$+$ clean\_mask} ($38.10$):
\begin{lstlisting}[style=pythoncode]
polygons = clean_mask(segment(bboxes))         # smooth ragged SAM boundaries
\end{lstlisting}
\noindent$\hookrightarrow$ \textbf{$+$ contextual query} ($39.20$):
\begin{lstlisting}[style=pythoncode]
bboxes = detect(f"{target} on {plant}", threshold=0.15)  # ground the disease in its plant
\end{lstlisting}
\noindent$\hookrightarrow$ \textbf{Final} (test Dice \textbf{39.38}):
\begin{lstlisting}[style=pythoncode]
bboxes   = detect(f"{target} on {plant}", threshold=0.15)
bboxes   = nms(filter_by_area(bboxes, 0.0002, 0.80), iou=0.55)  # loose NMS keeps co-located lesions
polygons = clean_mask(filter_polygons_by_area(segment(bboxes), 0.0001, 0.85))
\end{lstlisting}

\medskip
\noindent To show the observer side concretely, Listing~\ref{lst:analyzer_excerpt} reproduces a diagnostic program from a PlantSeg-OOD search run. It stratifies images by ground-truth lesion count, summarizes Dice and precision on the densest tier, and flags over-segmentation (predicted area exceeding twice the ground truth) --- the kind of plain-text observation the Producer consumes on its next revision.

\begin{lstlisting}[style=pythoncode,basicstyle=\ttfamily\footnotesize,caption={Analyzer program from the segmentation run: it profiles over- vs.\ under-detection from signed box error, flags small-false-positive floods, names the plant species the detector misses entirely, and detects the SAM mask-boundary ceiling (box Dice far exceeding mask Dice) --- the plain-text observations the Producer consumes on the next iteration.},label={lst:analyzer_excerpt}]
def analyze(per_image_results):
    observations = []
    # over- vs under-detection across the batch (signed box error)
    over  = [r for r in per_image_results if r['signed_bb_error'] > 0]
    under = [r for r in per_image_results if r['signed_bb_error'] < 0]
    zero  = [r for r in per_image_results if r['n_pred_bb'] == 0]
    mean_area_err = sum(r['signed_area_error'] for r in per_image_results) / len(per_image_results)
    observations.append(
        f"{len(over)} over-, {len(under)} under-detected, {len(zero)} with zero boxes; "
        f"mean area error {mean_area_err:+.3f} "
        f"({'masks over-shoot' if mean_area_err > 0 else 'masks under-cover'}).")

    # many boxes but little area -> small false-positive flood
    fp = [r for r in per_image_results
          if r['signed_bb_error'] > 2 and r['signed_area_error'] < 0]
    if fp:
        observations.append(f"{len(fp)} tasks flood small FPs -- tighten NMS or raise threshold.")

    # plant species the detector misses entirely
    if zero:
        plants = sorted(set(r['plant'] for r in zero))
        observations.append(f"zero detections on {plants} -- try multi-query or lower threshold.")

    # SAM boundary ceiling: box is right but mask Dice lags far behind
    gap = [r for r in per_image_results if r['dice_bbox'] - r['dice_mask'] > 0.3]
    if gap:
        observations.append(f"{len(gap)} tasks: bbox Dice >> mask Dice -- SAM boundary is the bottleneck.")
    return observations
\end{lstlisting}

\medskip\noindent
Both trajectories tell the same story through opposite strategies: from a one-line seed, the search turns \emph{specific, measured} analyzer diagnoses --- a named bottleneck image in counting, a mislocalising bare query in segmentation --- into targeted, auditable code revisions whose internal scores rise monotonically toward the deployed program. A one-shot synthesizer, which commits to a fixed pipeline without ever observing how it fails, has access to none of this.

\end{document}